\documentclass[runningheads]{llncs}
\usepackage{graphicx}
\usepackage{amsmath,amssymb} 
\usepackage{color}\usepackage[width=122mm,left=12mm,paperwidth=146mm,height=193mm,top=12mm,paperheight=217mm]{geometry}
\usepackage{url,xspace,caption,subcaption}
\usepackage{array,multirow,booktabs,balance}
\usepackage{times,txfonts}
\usepackage{hyperref}  
\hypersetup{backref,colorlinks=true,linkcolor=red,citecolor=green,urlcolor=blue}

\makeatletter
\DeclareRobustCommand\onedot{\futurelet\@let@token\@onedot}
\def\@onedot{\ifx\@let@token.\else.\null\fi\xspace}

\def\eg{\emph{e.g}\onedot} 
\def\ie{\emph{i.e}\onedot}

\def\etal{\emph{et al}\onedot}
\makeatother

\def\Vec#1{{\boldsymbol{#1}}}
\def\Mat#1{{\boldsymbol{#1}}}

\newcommand{\tr}{\mathop{\rm  Tr}\nolimits}

\begin{document}
\pagestyle{headings}
\mainmatter
\title{Kernel Coding: General Formulation and Special Cases}

\titlerunning{Kernel Coding: General Formulation and Special Cases}

\authorrunning{Harandi \etal}

\author
  {
  Mehrtash~T.~Harandi
  \and
  Mathieu~Salzmann
  }

\institute
  {
  Australian National University, Canberra, ACT 0200, Australia\\
  ~NICTA\thanks
    {
    NICTA is funded by the Australian Government as represented by the Department of
    Broadband, Communications and the Digital Economy, as well as by the Australian
    Research Council through the ICT Centre of Excellence program.
    }%
    , Locked Bag 8001, Canberra, ACT 2601, Australia%
  }

\maketitle

\begin{abstract}
Representing images by compact codes has proven beneficial for many visual recognition tasks. Most existing techniques, however, perform this coding step directly in image feature space, where the distributions of the different classes are typically entangled. In contrast, here, we study the problem of performing coding in a high-dimensional Hilbert space, where the classes are expected to be more easily separable. To this end, we introduce a general coding formulation that englobes the most popular techniques, such as bag of words, sparse coding and locality-based coding, and show how this formulation and its special cases can be kernelized. Importantly, we address several aspects of learning in our general formulation, such as kernel learning, dictionary learning and supervised kernel coding. Our experimental evaluation on several visual recognition tasks demonstrates the benefits of performing coding in Hilbert space, and in particular of jointly learning the kernel, the dictionary and the classifier.

\end{abstract}

\section{Introduction}
\label{sec:intro}

Over the years, \textbf{coding} -in its broadest definition-
has proven a crucial step in visual recognition systems~\cite{Boureau:CVPR:2010,Coates:ICML:2011}. 
Many techniques have been investigated, such as bag of words~\cite{Koenderink:IJCV:1999,Grauman:ICCV:2005,Lazebnik:CVPR:2006,Agarwal:ECCV:2006,%
VanGemert:PAMI:2010,Liu:ICCV:2011}, sparse coding~\cite{Olshausen:Nature:1996,Yang:CVPR:2009:SPM} and 
locality-based coding~\cite{Yu:NIPS:2009,Wang:CVPR:2010}. 
All these techniques follow a similar flow: Given a dictionary of codewords, a query is associated to one or multiple dictionary elements with different weights (\ie, binary or real). These weights, or {\it codes}, act as the new representation for the query and serve as input to a classifier (\eg, Support Vector Machine (SVM)) after an optional pooling step.

From the description above, it is quite clear that the quality of the codes will depend on the dictionary. Therefore, recent research has focused on learning codebooks that better reflect the underlying recognition problem from training data~\cite{Winn:ICCV:2005,Perronnin:PAMI:2008,Lazebnik:PAMI:2009}. 
In particular, supervised dictionary learning methods have been introduced to directly exploit the labels of the training samples in the objective function of the coding problem, thus jointly learning a codebook 
and a classifier~\cite{Mairal:NIPS:2009,Yang:CVPR:2010_Supervised,Lian:ECCV:2010,Jiang:PAMI:2013}.

In most existing techniques, coding is performed directly in image feature space. In realistic scenarios, the classes are not linearly separable in this space, and thus kernel-based classifiers are learned on the codes. However, the codes themselves may have been affected by the intricate distributions of the classes in feature space. Therefore, they may be sub-optimal for classification, even when employing a non-linear classifier. While supervised dictionary learning methods may help computing better-suited codes, they are currently limited to linear, or bilinear classifiers, which may be ill-suited for complex visual recognition tasks.

To alleviate the aforementioned issue, a handful of studies have considered performing coding in kernel space~\cite{Gao:ECCV:2010,Nguyen:TIP:2013}. It is widely acknowledged that mapping the data to a high-dimensional Hilbert space should make the classes more easily separable. Therefore, one can reasonably expect that codes extracted in Hilbert space should better reflect the classification problem at hand, while still yielding a compact representation of the data. However, as mentioned earlier, the literature on this topic remains limited, and exclusively focused on the specific case of sparse coding. In~\cite{Gao:ECCV:2010}, Gao \etal kernelized the Lasso problem and introduced a gradient descend approach to obtaining a dictionary in kernel space. This approach, however, is restricted to the Gaussian kernel. In~\cite{Nguyen:TIP:2013}, Van Nguyen \etal addressed the problem of unsupervised dictionary learning in kernel space by kernelizing the KSVD algorithm. While this generalizes kernel dictionary learning beyond the Gaussian kernel case, many questions remain unanswered. For instance:
\begin{enumerate}
\item \emph{Is there a principled way to find the best kernel space for the problem at hand?}
\item \emph{How can a supervised dictionary be learned in kernel space?}
\item \emph{Can coding schemes other than sparse coding be performed in the kernel space?}
\end{enumerate}

Here, we aim to answer such questions and therefore study several aspects of coding in a possibly infinite-dimensional Hilbert space. To this end, we first introduce a general formulation to the coding problem, which encompasses bags-of-words, sparse coding and locality-based coding. We show that, in all these cases, coding in a Hilbert space can be fully expressed in terms of kernels. 
We then show how the kernel, the dictionary, and the classifier can be learned in our general framework. This allows us to find the Hilbert space, codes and classifier that are jointly best-suited for the task at hand.

We evaluate our general kernel coding formulation on several image recognition benchmarks and provide a detailed analysis of its different special cases. Our empirical results evidence the strengths of our kernel coding approach over linear methods. Furthermore, it shows the benefits of jointly learning the kernel, dictionary and classifier over existing kernel sparse coding methods that only learn the codebook.

\section{Kernel Coding}
\label{sec:kernel_coding}

Given a query vector $\Vec{x} \in \mathbb{R}^d$, such as an image descriptor, many methods have been proposed to transform $\Vec{x}$ into a more compact, and hopefully discriminative, representation, hereafter referred to as {\it code}. 
Popular examples of such an approach include sparse coding and bag of words. In the following, we introduce a general formulation that englobes many different coding techniques, and lets us derive kernelized versions of these different methods.

More specifically, given a dictionary \mbox{$\Mat{D}_{d \times N} = \left[ \Vec{d}_1 | \Vec{d}_2| \cdots | \Vec{d}_N \right],\;\Vec{d}_i \in \mathbb{R}^d~$} containing $N$ elements, coding can be expressed as the solution to the optimization problem
\begin{align}
	&\underset{\Vec{y}}{\min}  \Bigl\| \Vec{x}- \sum\limits_{j=1}^{N} [\Vec{y}]_j \Vec{d}_j  \Bigr\|_2^2 + \gamma r( \Vec{y}; \Vec{x}, \Mat{D} ) \label{eqn:coding}\\
	&{\rm s.t.}~~~\Vec{y} \in \mathcal{C},\nonumber	
\end{align}
where $[\cdot]_j$ denotes the $j^{th}$ element of a vector, $r(\cdot)$ is a prior on the codes $\Vec{y}$, and $\mathcal{C}$ is a set of constraints on $\Vec{y}$. Note that this formulation allows the prior to be dependent on both the query $\Vec{x}$ and the dictionary $\Mat{D}$. Although not explicit, this is also true for $\mathcal{C}$.

Our goal here is to perform nonlinear coding in the hope to obtain a better representation. To this end, let $\phi:\mathbb{R}^d \to \mathcal{H}$ be a mapping to a Reproducing Kernel Hilbert Space (RKHS) induced by the kernel $k(\Vec{x}_i,\Vec{x}_j) = \phi(\Vec{x}_i)^T\phi(\Vec{x}_j)$. Then, coding in $\mathcal{H}$ can be formulated as
\begin{align}
	&\underset{\Vec{y}}{\min}  \Bigl\| \phi\big(\Vec{x})- \sum\limits_{j=1}^{N} [\Vec{y}]_j \phi\big(\Vec{d}_j)  \Bigr\|_2^2 + \gamma r( \Vec{y}; \phi\big(\Vec{x}), \phi\big(\Mat{D}) ) \label{eqn:nonlin_coding}\\
	&{\rm s.t.}~~~\Vec{y} \in \mathcal{C}.\nonumber	
\end{align}
Expanding the first term in~\eqref{eqn:nonlin_coding} yields
\noindent
\begin{align}
    \hspace{-0.5cm}\Big\| \phi(\Vec{x})-\sum\limits_{j=1}^{N}[\Vec{y}]_j \phi(\Vec{d}_j) \Big\|_2^2 
    &=   \phi(\Vec{x})^T\phi(\Vec{x}) 
    -2\sum\limits_{j=1}^{N}[\Vec{y}]_j \phi(\Vec{d}_j)^T \phi(\Vec{x})
    +\sum\limits_{i,j=1}^{N}[\Vec{y}]_i [\Vec{y}]_j\phi(\Vec{d}_i)^T\phi(\Vec{d}_j)\nonumber \\  
    & = k(\Vec{x},\Vec{x})-2\sum\limits_{j=1}^{N}{[\Vec{y}]_j k(\Vec{x},\Vec{d}_j)} +
    \sum\limits_{i,j=1}^{N}{[\Vec{y}]_i [\Vec{y}]_j k(\Vec{d}_i,\Vec{d}_j)} \nonumber \\
    & = k(\Vec{x},\Vec{x})-2\Vec{y}^T\Vec{k}(\Vec{x},\Mat{D})+
    \Vec{y}^T \Mat{K}(\Mat{D},\Mat{D}) \Vec{y},
    \label{eqn:KSR_Opt}
\end{align}%
\noindent
where $\Vec{k}(\Vec{x},\Mat{D}) \in \mathbb{R}^{N \times 1}$, and $\Mat{K}(\Mat{D},\Mat{D}) \in \mathbb{R}^{N \times N}$. This shows that the reconstruction term in~\eqref{eqn:coding}, common to most coding techniques, can be kernelized. More importantly, after kernelization, this term remains quadratic, convex and similar to its counterpart in Euclidean space.
In the remainder of this section, we discuss the priors $r(\cdot)$ and constraint sets $\mathcal{C}$ corresponding to specific coding techniques, and show how they can also be kernelized.

\subsection{Kernel Bag of Words}

Bag of words~\cite{Salton:1988} is possibly the simplest coding technique, which consists of assigning the query $\Vec{x}$ to a single dictionary element. In our framework, this can be expressed by having no prior $r(\cdot)$ and defining the constraint set $\mathcal{C}$ as
\begin{equation}
\mathcal{C} = \left\{ \Vec{y}\mid \Vec{y} \in \left\{0,1\right\}^N, \sum_i \Vec{y}_i = 1 \right\}\;.
\end{equation}
Note that this constraint set only depends on the code $\Vec{y}$. Therefore, kernelizing this model does not require any other operation than that described in Eq.~\ref{eqn:KSR_Opt}. In other words, $\Vec{y}$ can be obtained by finding the nearest neighbor to $\Vec{x}$ in kernel space and setting the corresponding entry in $\Vec{y}$ to 1.

The above-mentioned hard assignment bag of words model was relaxed to soft assignments~\cite{VanGemert:PAMI:2010}, where the query is assigned to multiple dictionary elements, with weights depending on the distance between these elements and the query. In this soft assignment bag of words model, a code $\Vec{y}_i$ is directly constrained to take the value
\begin{equation}
\Vec{y}_i = \frac{\exp\left(-\sigma \|\Vec{x} - \Vec{d}_i\|^2_2\right)}{\sum_j \exp\left(-\sigma \|\Vec{x} - \Vec{d}_j\|^2_2\right)}\;.
\end{equation}
Here, $\sigma > 0$ is the bandwidth parameter of the algorithm. To kernelize this model, we note that each exponential term can be written as
\begin{equation}
\exp\left(-\sigma \|\phi(\Vec{x}) - \phi(\Vec{d}_i)\|^2_2\right) = \exp\left(-\sigma \left( k(\Vec{x},\Vec{x}) - 2k(\Vec{x},\Vec{d}_i) + k(\Vec{d}_i,\Vec{d}_i)\right)\right) \;. \label{eqn:kernel_soft_assign}
\end{equation}
Since this only depends on kernel values, so do the resulting codes $\Vec{y}_i$. In practice, soft assignment bags of words have often proven more effective than hard assignment ones, especially when a single code is used to represent an entire image.

\subsection{Kernel Sparse Coding}

Over the years, sparse coding has proven effective at producing compact and discriminative representations of images~\cite{ELAD_SR_BOOK_2010}. From~\eqref{eqn:coding}, sparse coding can be obtained by employing the prior
\begin{equation}
r(\Vec{y}) = \|\Vec{y}\|_1\;,
\end{equation}
which corresponds to a convex relaxation of the $\ell_0$-norm encoding sparsity, and not using any constraints. Since this prior only depends on $\Vec{y}$, the only step required to kernelize sparse coding is given in Eq.~\ref{eqn:KSR_Opt}. Note that any structured, or group, sparsity prior can also be utilized in the same manner.

A solution to kernel sparse coding can be obtained by transforming the resulting optimization problem into a standard vectorized sparse coding problem. To this end, let $\Mat{U}\Mat{\Sigma}\Mat{U}^T$ be the SVD of $\Mat{K}(\Mat{D},\Mat{D})$. Then it can easily be verified that kernel sparse coding is equivalent to the optimization problem
\begin{equation}
	\underset{\Vec{y}}{\min} \:
	\Vert \Vec{\tilde{x}} - \Mat{A}\Vec{y}\Vert^2_2 +\gamma \|\Vec{y}\|_1, 
	\label{eqn:Opt_kernel4}
\end{equation} 
where $\Mat{A} = \Mat{U}\Mat{\Sigma}^{1/2}$ and $\Vec{\tilde{x}} = \Mat{U}\Mat{\Sigma}^{-1/2}\Vec{k}(\Vec{x},\Mat{D})$. As a consequence, any efficient sparse solver such as SLEP~\cite{Liu:2009:SLEP:manual} or SPAMS~\cite{SPAMS:JMLR} can be employed to solve kernel sparse coding.

\subsection{Locality-Constrained Kernel Coding}

As with bags of words, the notion of locality between the query and the dictionary elements has also been employed in the sparse coding framework. This was first introduced in the Local Coordinate Coding (LCC) model~\cite{Yu:NIPS:2009}, which was then modified as the Locality-Constrained Linear Coding (LLC) model~\cite{Wang:CVPR:2010}, whose solution can be obtained more efficiently. We therefore focus on kernelizing the LLC formulation, though it can easily be verified that LCC can be kernelized in a similar manner.
From our general formulation~\eqref{eqn:coding}, LLC can be obtained by defining
\begin{equation}
r(\Vec{y};\Vec{x},\Vec{D}) = \left\|\Mat{E}\Vec{y}\right\|_2^2,\;{\rm with} \; \Mat{E}_{ii} = \exp\left(\sigma\|\Vec{x} - \Vec{d}_i\|_2\right), \; {\rm and} \;\; \Mat{E}_{ij} = 0, \; i \neq j\;,
\end{equation}
and $\mathcal{C} = \{\Vec{y}\mid \sum_i \Vec{y}_i = 1\}$. Since the constraints only depend on the codes, they do not require any special care for kernelization. To kernelize the prior, we note that
\begin{equation}
\small{
\Mat{E}_{ii} = \exp\left(-\sigma \left\|\phi(\Vec{x}) - \phi(\Vec{d}_i)\right\|_2\right) = \exp\left(-\sigma\sqrt{k(\Vec{x},\Vec{x}) - 2k(\Vec{x},\Vec{d}_i) + k(\Vec{d}_i,\Vec{d}_i)}\right).
}
\end{equation}
Therefore, $\Mat{E}$ can be expressed solely in terms of kernel values, and so does $r(\Vec{y};\Vec{x},\Vec{D})$.

In~\cite{Wang:CVPR:2010}, the initial LLC formulation was then approximated to improve the speed of coding. To this end, for each query, the dictionary $\Mat{D}$ was replaced by a local dictionary $\Mat{B}$ formed by the $N_\Mat{B}$ nearest dictionary elements to $\Vec{x}$. In Hilbert space, we can follow a similar idea and obtain a local dictionary $\Mat{B}$ by performing kernel nearest neighbor between the dictionary elements and the query. This lets us write LLC in Hilbert space as
\begin{align}
	&\underset{\Vec{y}}{\min} \Vert \phi(\Vec{x}) -  \phi(\Mat{B})\Vec{y}  \Vert^2_2 
	\label{eqn:kllc_approx}\\
	&{\rm s.t.}~~~\Vec{1}^T\Vec{y} = 1,\nonumber	
\end{align}
which has a form similar to~\eqref{eqn:nonlin_coding} with no prior. This can then be directly kernelized by making use of Eq.~\ref{eqn:KSR_Opt}.

The solution to kernel LLC can be obtained from the Lagrange dual~\cite{Boyd:2004} of Eq.~\eqref{eqn:kllc_approx}, which can be written as
\begin{equation}
L_{kllc}(\Vec{y},\lambda) = k(\Vec{x},\Vec{x})-2\Vec{y}^T\Vec{k}(\Vec{x},\Mat{D})+ \Vec{y}^T \Mat{K}(\Mat{D},\Mat{D}) \Vec{y} + \lambda (\Vec{y}^T\Vec{1} - 1) \;.
\end{equation}
The gradient of $L_{kllc}(\Vec{y},\lambda)$ w.r.t. $\Vec{y}$ can then be expressed as
\begin{align}
	\nabla_\Vec{y}L_{kllc} &= -2\Vec{k}(\Vec{x},\Mat{D})+ 2\Mat{K}(\Mat{D},\Mat{D}) \Vec{y} + \lambda \Vec{1} \notag\\
	&= -2\sum\nolimits_j\Vec{y}_j\Vec{k}(\Vec{x},\Mat{D})+ 2\Mat{K}(\Mat{D},\Mat{D}) \Vec{y} + \lambda \Vec{1} \notag\\
	&= -2\left(\Vec{1}^T \otimes \Vec{k}(\Vec{x},\Mat{D})\right)\Vec{y}+ 2\Mat{K}(\Mat{D},\Mat{D}) \Vec{y}  + \lambda \Vec{1} \;,
\end{align}
where the second line was obtained by making use of the constraint on $\Vec{y}$. By setting this gradient to $0$, $\Vec{y}$ can be obtained as the solution to the linear system $\big(\Mat{K}(\Mat{D},\Mat{D}) - ( \Vec{1}^T \otimes \Vec{k}(\Vec{x},\Mat{D}))\big)\Vec{y} = \Vec{1}$, and then normalized by its $\ell_1$ norm to satisfy the constraint\footnote{Note that the dependency on $\lambda$ is ignored in the linear system since it would only influence the global scale of $\Vec{y}$, which would then be canceled out by the normalization.}.

\section{Learning in Kernel Coding}

In this section, we discuss the learning of the different components of the kernel coding methods described in Section~\ref{sec:kernel_coding}. In particular, given a set of $M$ training samples $\mathcal{X} = \{\Vec{x}_i\}_{i=1}^M$, we investigate how the kernel, as well as the dictionary can be learned. Furthermore, we show how a discriminative classifier can be directly introduced and learned within our kernel coding framework. While, in the following, we consider the different learning problems separately, they can of course be solved jointly by employing the alternating minimization approach commonly used in dictionary learning.

\subsection{Kernel Learning}

We first tackle the problem of finding the best kernel for the problem at hand. While the kernel could, in principle, be considered as a completely free entity, this typically yields an under-constrained formulation and only suits the transductive settings. Here instead, we assume that the kernel has either a parametric form (\eg, Gaussian kernel with width as a parameter), or can be represented as a linear combination of fixed kernels (\ie, multiple kernel learning).

\subsubsection{Learning Kernel Parameters}\mbox{}\\
\label{subsec:learning_kernel_parameter}

Let us assume that we are given a fixed dictionary $\Mat{D}$, and that our kernel function has a parametric form with parameters $\beta$. Following our general formulation, $\beta$ can be learned by solving the optimization problem
\begin{align}
	&\underset{\beta,\{\Vec{y}_i\}}{\min} \frac{1}{M}\sum_{i=1}^{M} L_{\phi}(\beta,\Vec{y}_i; \Vec{x}_i,\Mat{D}) \label{eqn:learn_beta}\\
	&{\rm s.t.}~~~\Vec{y}_i \in \mathcal{C},\; \forall i \in [1,M], \nonumber	
\end{align}
where $\Vec{y}_i$ is the vector of sparse codes for the $i^{th}$ training sample $\Vec{x}_i$, and $L_{\phi}(\cdot)$ is the kernelized 
objective function defined in~\eqref{eqn:nonlin_coding}.

Note that~\eqref{eqn:learn_beta} is not jointly convex in $\beta$ and $\{\Vec{y}_i\}_{i=1}^M$. Therefore, we follow the standard alternating minimization strategy that consists of iteratively fixing one variable (\ie, either $\beta$, or the $\Vec{y}_i$s) and solving for the other. In general, we cannot expect the objective function to be convex in $\beta$, even with fixed $\{\Vec{y}_i\}$. Therefore, we utilize a gradient-based trust-region method to obtain $\beta$ at each iteration.

While any kernel function could be employed in this framework, in practice, we make use of the Gaussian kernel 
$k(\Vec{x}_i,\Vec{x}_j) = \exp(-\beta \|\Vec{x}_i,\Vec{x}_j\|^2_2)$. Unfortunately, with this kernel,~\eqref{eqn:learn_beta} is ill-posed in terms of $\beta$. More specifically, $\beta = 0$ is a minimum of~\eqref{eqn:learn_beta}. This is due to the fact that, if $\beta \to 0$, all samples in the induced Hilbert space $\mathcal{H}$ collapse to one point, \ie, $\|\phi(\Vec{x}_i) - \phi(\Vec{x}_j)\|^2 = k(\Vec{x}_i,\Vec{x}_i) - 2k(\Vec{x}_i,\Vec{x}_j) - k(\Vec{x}^j,\Vec{x}^j) = 0$\footnote{The same statement holds for polynomial kernels of the form $k(\Vec{x}_i,\Vec{x}_j) = (1+\beta \Vec{x}_i^T\Vec{x}_j)^p$.}. 

To avoid this trivial solution, we propose to search for a $\beta$ that not only minimizes the kernel coding cost, but also maximizes a measure of discrepancy between the dictionary atoms in $\mathcal{H}$. In other words, we search for a Hilbert space $\mathcal{H}$ that simultaneously yields a diverse dictionary and a good representation of the data. To this end, we define the discrepancy between the dictionary atoms in 
$\mathcal{H}$ as
\begin{equation}
	J_\phi(\Mat{D},\beta) = \frac{1}{N^2}\hspace{-0.2cm}\sum\limits_{(i,j) = 1}^{N}\hspace{-0.2cm}\|\phi(\Vec{d}_i) - \phi(\Vec{d}_j)\|^2_2 = \frac{1}{N^2}\hspace{-0.2cm}\sum\limits_{(i,j) = 1}^{N}\hspace{-0.2cm}\left( k(\Vec{d}_i,\Vec{d}_i) - 2k(\Vec{d_i},\Vec{d}_j) + k(\Vec{d}_j,\Vec{d}_j) \right). 
	\label{eqn:discrepancy_atoms}
\end{equation} 
Given $\{\Vec{y}_i\}$, this lets us obtain $\beta$ by solving the optimization problem
\begin{align}
	&\underset{\beta}{\min} \frac{\frac{1}{M}\sum_{i=1}^{M} L_{\phi}(\beta,\Vec{y}_i; \Vec{x}_i,\Mat{D})}{J_\phi(\Mat{D},\beta)}\;. \label{eqn:learn_beta_ratio} \\
	&\mathrm{s.t.}~~~\Vec{y}_i \in \mathcal{C},\; \forall i \in [1,M] \nonumber	
\end{align}
We obtain a local minimum to this problem using a gradient-based trust-region method.
Note that, in most of the special cases discussed in Section~\ref{sec:kernel_coding}, the prior $r(\cdot)$ does not depend on the kernel and can thus be omitted when updating $\beta$. This is also the case of the constraint set $\mathcal{C}$, except for soft assignment kernel bag of words where the constraints directly define the codes. In this case, however, the codes can be replaced by the form in 
Eq.~\ref{eqn:kernel_soft_assign} in the objective function, which then becomes a function of $\beta$ solely.

\subsubsection{Multiple Kernel Coding} \mbox{}\\
\label{subsec:mkl_SC}

In various applications, combining multiple kernels has often proven more effective than using a single kernel~\cite{Bucak:PAMI:2013}. 
Following this idea, we propose to model our kernel $k(\cdot,\cdot)$ as a linear combination of a set of kernels, \ie, $k(\Vec{x}_i,\Vec{x}_j) = \sum_{l=1}^L \alpha_l k_l(\Vec{x}_i,\Vec{x}_j)$. Kernel learning then boils down to finding the best coefficients $\Vec{\alpha}$. In our context, this can be done by solving the optimization problem
\begin{align}
	&\underset{\Vec{\alpha},\{\Vec{y}_i\}}{\min} \frac{\frac{1}{M}\sum_{i=1}^{M} L_{\phi}(\Vec{\alpha},\Vec{y}_i; \Vec{x}_i,\Mat{D})}{J_\phi(\Mat{D},\Vec{\alpha})}\;, \label{eqn:learn_alpha_ratio} \\
	&{\rm s.t.}~~~\Vec{y}_i \in \mathcal{C},\; \forall i \in [1,M], \nonumber	
\end{align}
where $L_{\phi}(\cdot)$ and $J_{\phi}(\cdot)$ are defined as before, but in terms of a kernel given as the linear combination of multiple base kernels with weights $\Vec{\alpha}$. As in the parametric case, we obtain $\Vec{\alpha}$ using a gradient-based trust-region method.

\subsection{Dictionary Learning}
\label{sec:dic_learning}

In many cases, it is beneficial to not only learn codes for a given dictionary, but optimize the dictionary to best suit the problem at hand. Here, we show how this can be done in our general formulation.
Given fixed kernel parameters and codes for the training data, learning the dictionary can be expressed as solving the optimization problem
\begin{align}
	&\underset{\Mat{D}}{\min} \frac{1}{M}\sum_{i=1}^{M} L_{\phi}(\Mat{D}; \Vec{y}_i, \Vec{x}_i)\;. \label{eqn:learn_dic} \\
	&{\rm s.t.}~~~\Vec{y}_i \in \mathcal{C},\; \forall i \in [1,M]. \nonumber	
\end{align}
Here, we make use of the \emph{Representer theorem}~\cite{Scholkopf:COLT:2001} which enables us to express the dictionary as a linear combination of the training samples in RKHS. That is
\begin{equation}
	\phi(\Vec{d}_r) = \sum\limits_{i = 1}^{M} a_{r,i}\phi(\Vec{x}_i),
\end{equation}
where $\{a_{r,i}\}$ is the set of weights, now corresponding to our new unknowns. By stacking these weights for the $N$ dictionary elements and the $M$ training samples in a matrix $\Mat{A}_{M \times N} = [\Vec{a}_{1}|\Vec{a}_{2}|\cdots|\Vec{a}_N]$, with $\Vec{a}_r = [a_{r,1},a_{r,2},\cdots,a_{r,M}]^T$, we have
\begin{equation}
	\Mat{\Phi}(\Mat{D}) = \Mat{\Phi}(\Mat{X})\Mat{A}\;.
\end{equation}

For most of the algorithms presented in Section~\ref{sec:kernel_coding}, \ie, kernel bag of words, kernel sparse coding and approximate kernel LLC, the only term that depends on the dictionary is the reconstruction error (the first term in the objective of~\eqref{eqn:nonlin_coding}). Given the matrix of sparse codes $\Mat{Y}_{N \times M}$, this term can be expressed as
\begin{align}
	R(\Mat{A}) &=  \big\|\Mat{\Phi}(\Mat{X}) - \Mat{\Phi}(\Mat{X})\Mat{A}\Mat{Y} \big\|_F^2 \notag \\
	&= \tr \left( \Mat{\Phi}(\Mat{X}) (\mathbf{I}_M - \Mat{A}\Mat{Y})(\mathbf{I}_m - \Mat{A}\Mat{Y})^T \Mat{\Phi}(\Mat{X})^T\right)\notag \\
	&= \tr \left( \Mat{K}(\Mat{X},\Mat{X}) (\mathbf{I}_M - \Mat{A}\Mat{Y} - \Mat{Y}^T\Mat{A}^T + 
	\Mat{A}\Mat{Y}\Mat{Y}^T\Mat{A}^T) \right)\;.\notag \\
	\label{eqn:dic_learn_cost_func}
\end{align}
The new dictionary can then be obtained by zeroing out the gradient of this term w.r.t. $\Mat{A}$. This yields
\begin{align}
	\nabla{R}(\Mat{A}) = 0 &\Rightarrow 2\Mat{Y} = 2\Mat{Y}\Mat{Y}^T\Mat{A} \notag \\
	&\Rightarrow \Mat{A} = (\Mat{Y}\Mat{Y}^T)^{-1}\Mat{Y} = \Mat{Y}^\dagger \;.
	\label{eqn:gradient_dic_learn_cost_func}
\end{align}
In the case of approximate kernel LLC, we update the full dictionary at once. To this end, for each training sample $i$, we simply set to 0 the codes corresponding to the elements that do not belong to the local dictionary $\Mat{B}_i$. For soft-assignment kernel bag of words, where the constraints depend on the dictionary, the update of Eq.~\ref{eqn:gradient_dic_learn_cost_func} is not valid and one must resort to an iterative gradient-based update of $\Mat{A}$.

\subsection{Supervised Kernel Coding}
\label{sec:supervised_coding}
In the context of sparse coding, several works have studied the idea of making the sparse codes discriminative, and thus jointly learn a classifier with the codes and the dictionary~\cite{Mairal:NIPS:2009,Jiang:PAMI:2013}. Here, we introduce two formulations that also make use of supervised data in our general kernel coding framework. To this end, given data belonging to $S$ different classes, let $\Vec{l}_i$ be the $S$-dimensional binary vector encoding the label of training sample $\Vec{x}_i$, \ie, $[\Vec{l}_i]_j = 1$ is sample $i$ belongs to class $j$.

In our first formulation, we make use of a linear classifier acting on the codes. The prediction of such a classifier takes the form $\hat{\Vec{l}} = \Mat{W}\Vec{y}$. By employing the square loss, learning in Hilbert space can then be written as
\begin{align}
	&\underset{\Mat{W},\Mat{D},\{\Vec{y}_i\}}{\min} \frac{1}{M}\sum_{i=1}^{M} L_{\phi}(\Mat{D},\Vec{y}_i; \Vec{x}_i) + \frac{\eta}{M}\sum_{i=1}^M\left\|\Vec{l}_i - \Mat{W}\Vec{y}_i\right\|_2^2 + \rho\|\Mat{W}\|_F^2\;, \label{eqn:learn_supervised} \\
	&{\rm s.t.}~~~\Vec{y}_i \in \mathcal{C},\; \forall i \in [1,M], \nonumber	
\end{align}
where we utilize a simple regularizer on the parameters $\Mat{W}$. Following an alternating minimization approach, given $\Mat{D}$ and 
$\{\Vec{y}_i\}$, the classifier parameters $\Mat{W}$ can be obtained by solving a linear system arising from zeroing out the gradient of the second and third terms in the objective function. While the update for the dictionary is unaffected by the discriminative term, computing the codes must be modified accordingly. However, the least-squares form of this term makes it easy to introduce into the solutions of kernel sparse coding and approximate kernel LLC. 

For our second formulation, we employ a bilinear classifier, as suggested in~\cite{Mairal:NIPS:2009}. For a single class $j$, the prediction of this classifier is given by $\hat{\Vec{l}}_j = \phi(\Vec{x}^i)^T\Mat{W}_j\Vec{y}^i$, which requires one parameter matrix per class. Using the same loss as before, learning can then be expressed as
\begin{align}
	&\hspace{-0.3cm}\underset{\{\Mat{W}_j\},\Mat{D},\{\Vec{y}_i\}}{\min} \frac{1}{M}\sum_{i=1}^{M} L_{\phi}(\Mat{D},\Vec{y}_i; \Vec{x}_i) + \frac{\eta}{M}\sum_{i=1}^M\sum_{j=1}^S\left([\Vec{l}_i]_j - \phi(\Vec{x}_i)^T\Mat{W}_j\Vec{y}_i\right)^2 + \frac{\rho}{S}\sum_{j=1}^S\|\Mat{W}_j\|_F^2 \label{eqn:learn_supervised_bilinear} \\
	&{\rm s.t.}~~~\Vec{y}_i \in \mathcal{S},\; \forall i \in [1,M]. \nonumber	
\end{align}
By making use of the Representer theorem, we can express each parameter matrix $\Mat{W}_j$ as a linear combination of the training samples in Hilbert space, which yields
\begin{equation}
\Mat{W}_j = \Phi(\Mat{X})\Mat{A}_j\;.
\end{equation}
With this form, the prediction of the classifier is now given by
\begin{equation}
\hat{\Vec{l}_j} = \phi(\Vec{x}_i)^T\Phi(\Mat{X})\Mat{A}_j\Vec{y}_i = \Vec{k}(\Vec{x}_i,\Mat{X})\Mat{A}_j\Vec{y}_i \;,
\end{equation}
which depends on the kernel function. Similarly, it can easily be checked that the regularizer on the parameters $\Mat{W}_j$ can be expressed in terms of kernel values. Learning can then be formulated as a function of the matrices $\{\Mat{A}_j\}$, which appear in two convex quadratic terms, and can thus be computed as the solution to a linear system obtained by zeroing out the gradient of the objective function. Similarly to our first formulation, the dictionary update is unaffected by the discriminative terms, and the computation of the codes must account for the additional quadratic terms.

For both formulations, at test time, given a new sample $\Vec{x}^*$, we first compute the codes $\Vec{y}^*$ using the chosen coding approach from Section~\ref{sec:kernel_coding}, and then predict the label of the sample by applying the learned classifier.

\section{Experimental Evaluation}
\label{sec:experiments}

In this section, we demonstrate the strength of kernel coding on the tasks of material categorization, scene classification, and object and face recognition. We also demonstrate the ability of kernel coding to performing recognition on a Riemannian manifold whose non-linear geometry makes linear coding techniques inapplicable. 

Throughout our experiments, we refer to the different coding techniques as 
\begin{itemize}
	\item \textbf{kBOW}: kernel (soft-assignment) bag of words,
	\item \textbf{kSC}: kernel sparse coding~\eqref{eqn:Opt_kernel4},
	\item \textbf{kLLC}: (approximate) kernel LLC~\eqref{eqn:kllc_approx}, 
	\item \textbf{l-SkSC}: supervised kernel sparse coding with the linear classifier from~\eqref{eqn:learn_supervised},
	\item \textbf{l-SkLLC}: supervised (approximate) kernel LLC with the linear classifier from~\eqref{eqn:learn_supervised},
	\item \textbf{bi-SkSC}: supervised kernel sparse coding with the bilinear classifier from~\eqref{eqn:learn_supervised_bilinear}, and
	\item  \textbf{bi-SkLLC}: supervised (approximate) kernel LLC with the bilinear classifier from~\eqref{eqn:learn_supervised_bilinear}.
\end{itemize}

When not learning a dictionary, classification is performed as proposed in~\cite{Wright:PAMI:2009}. More specifically, given a query $\Vec{x}$ and its code $\Vec{y}$, let  
{$\Vec{y}_{(s)} = ( [\Vec{y}]_1\delta( \gamma_1-s ), ~[\Vec{y}]_2\delta( \gamma_2 - s ), ~\cdots, ~[\Vec{y}]_N\delta( \gamma_N-s ) )^T$}
be the class-specific sparse codes for class $s$, where {$\gamma_j \in \mathcal{S}$} is the class label of atom~$\Vec{d}_j$
and $\delta(\cdot)$ is the discrete Dirac function. The residual error of query $\Vec{x}$ for class $s$ can be defined as
\begin{align}
    \varepsilon_s(\Vec{x}) &=  \Big\| \Mat{\phi(\Vec{x})} - \sum\limits_{j=1}^{N} [\Vec{y}]_j \phi(\Vec{d}_j) \delta( \gamma_j-s ) \Big\|^2 = -2\Vec{y}_{(s)}^T\Vec{k}(\Vec{x},\Mat{D})+\Vec{y}_{(s)}^T \Mat{K}(\Mat{D},\Mat{D}) \Vec{y}_{(s)}\;,
    \label{eqn:sparse_residual_error}
\end{align}%
where the term $k(\Vec{x},\Vec{x})$ was dropped since it does not depend on the class label. The label of $\Vec{x}$ is then chosen as the class with minimum residual error $\varepsilon_s(\Vec{x})$.
In the case of unsupervised dictionary learning, since the dictionary elements do not have any associated labels anymore, we make use of a simple nearest-neighbor classifier based on the learned codes. Finally, in the supervised scenario, we first obtain the codes and then employ the learned classifier to obtain the labels.

In the following, we used the Gaussian kernel as base kernel, unless stated otherwise. The width of the Gaussian was learned using the method described in Section~\ref{subsec:learning_kernel_parameter} except for the multiple kernel learning experiments. We emphasize that, throughout  this section, coding is performed at image level, as in, \eg,~\cite{Jiang:PAMI:2013}. In other words, we represent each image with one descriptor to which a coding technique is applied.

\subsection{Flicker Material}

As a first experiment, we tackled the problem of material recognition using the Flicker Material Database (FMD)~\cite{FM_DATABASE}, which contains ten categories of materials (\ie, \textit{fabric}, \textit{foliage}, \textit{glass}, \textit{leather},\textit{metal}, \textit{paper}, \textit{plastic}, \textit{stone}, \textit{water} and \textit{wood}). 
Each category comprises 100 images, split into 50 close-up views and 50 object-scale views (see Figure.~\ref{fig:FMD_Dataset}). 
To describe each image, we used a bag of words model with 800 atoms computed from RootSIFT features~\cite{Arandjelovic:CVPR:2012} sampled every 4 pixels. 
Following the protocol used in~\cite{Sharan:IJCV:2013}, half of the images from each category was randomly chosen for training and the rest was used for testing. We report the average classification accuracy over 10 such random partitions. 

We first compare the performance of kBOW, kSC, kLLC and linear coding techniques (SC and LLC) without dictionary learning. To this end, we employed all the training samples as dictionary atoms.
In Table~\ref{tab:table_FMD_performance1}, we report the performance of the different methods on FMD. Note that, with the exception of kBOW, kernel coding techniques significantly outperform the linear ones. In particular, the gap between LLC and its kernel version reaches $4\%$.

Using the same data, and still without learning a dictionary, we evaluated the performance of multiple kernel coding as described in Section~\ref{subsec:mkl_SC}.
To this end, we combined 4 different kernels: a linear kernel, a Gaussian kernel, a second order polynomial kernel and a Sigmoid kernel. After learning the combination, kSC and kLLC achieved $\bf{59.1\%}$ and $\bf{58.5\%}$, respectively. This shows that combining kernels can even further improve the results of kernel coding techniques. Note that, in this manner, multiple kernel coding also allows us to nicely combine different features.

\begin{figure}[!tb]
	\begin{minipage}{0.45 \textwidth}\centering
		\includegraphics[width = \textwidth]{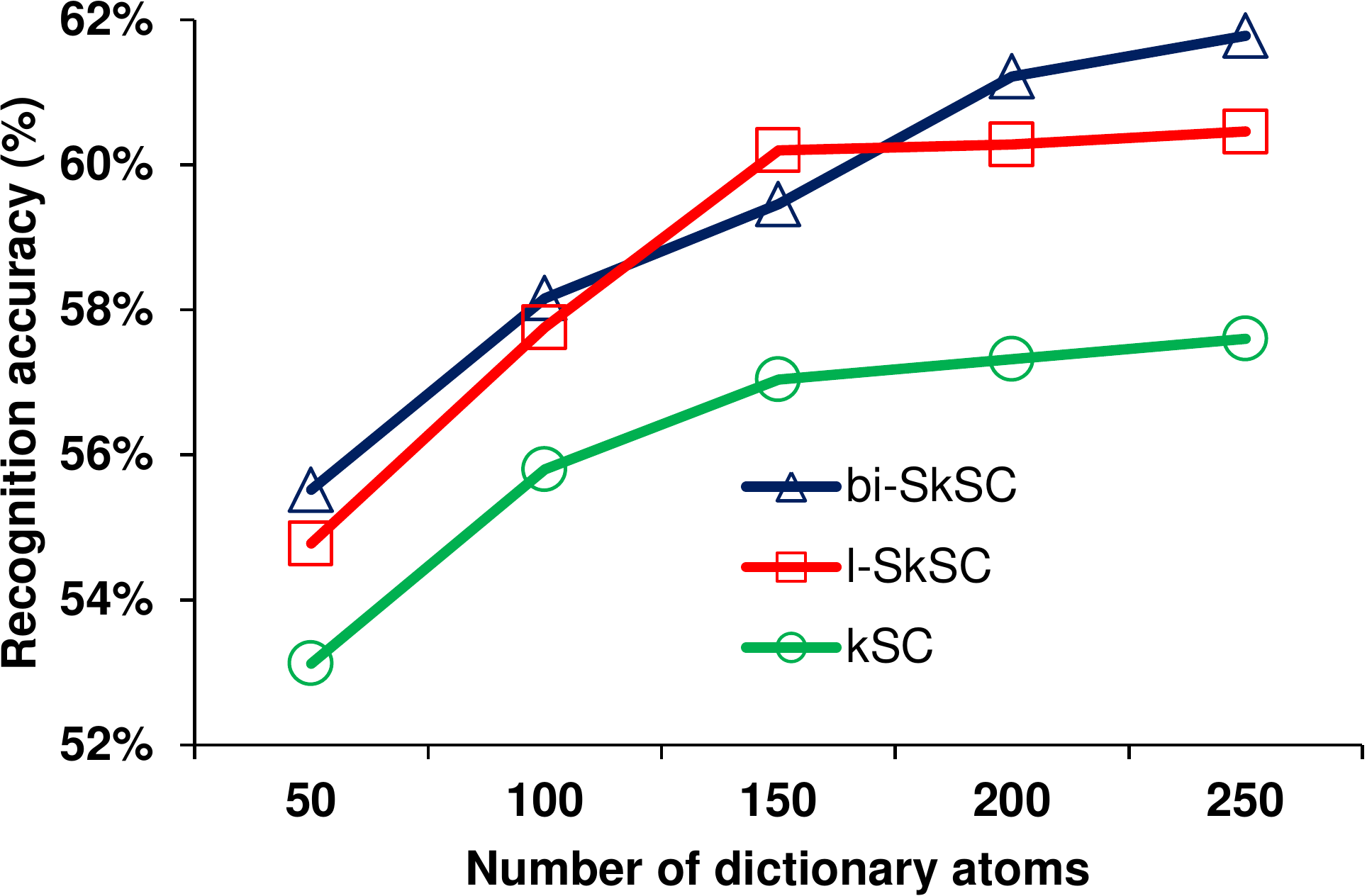}
	\end{minipage}
	\hfill
	\begin{minipage}{0.45 \textwidth}\centering
		\includegraphics[width = \textwidth]{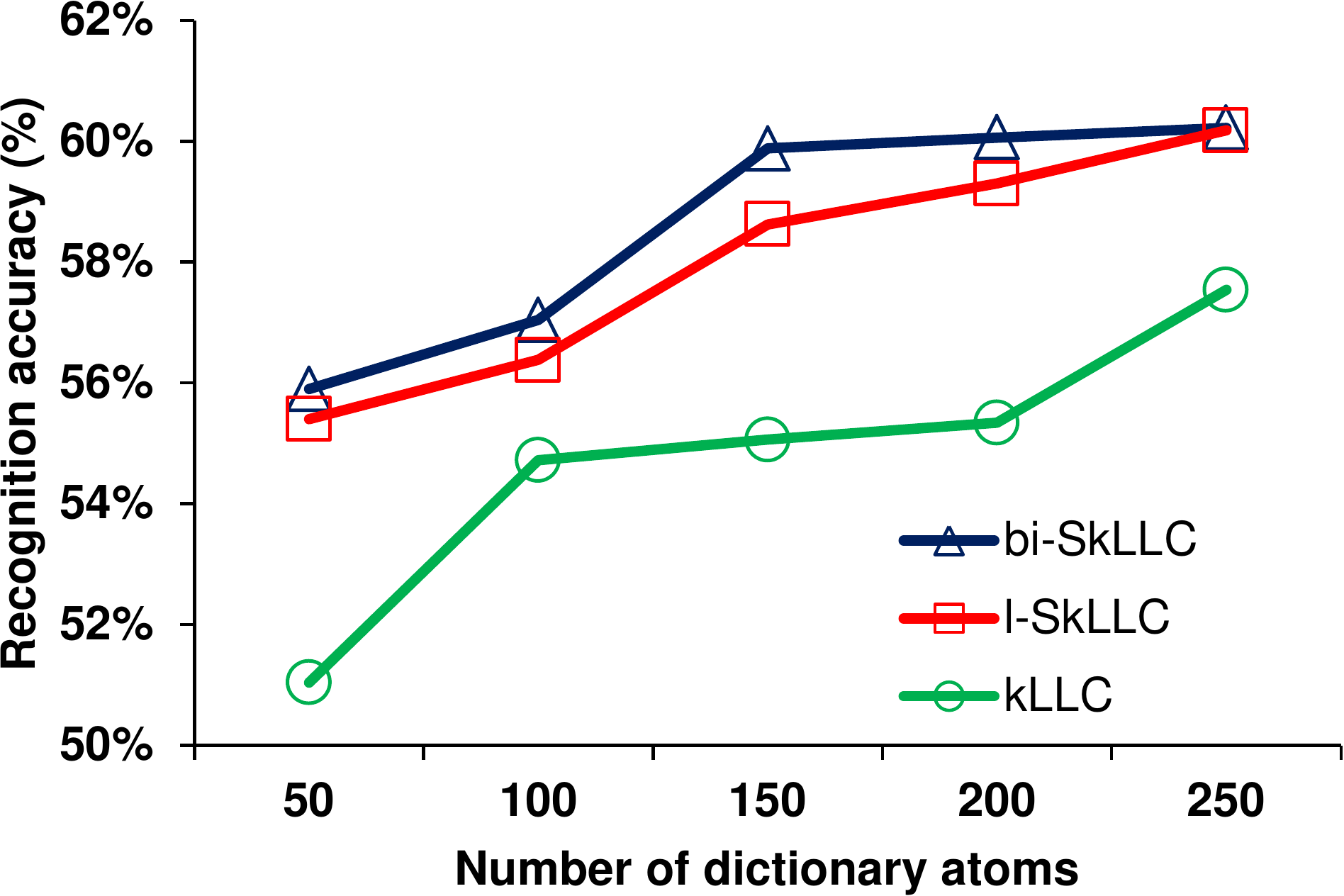}
	\end{minipage}			
	\vspace{-0.2cm}
	\caption{\footnotesize Recognition accuracies of dictionary learning for unsupervised and supervised kernel coding on FMD.}
	\label{fig:FMD_kSC_dic}
	\vspace{-0.4cm}
\end{figure}

We then turn to the cases of dictionary learning for unsupervised and supervised kernel coding. In the unsupervised cases, the dictionary was obtained using the method described in Section~\ref{sec:dic_learning}. For the supervised methods, we made us of the technique described in Section~\ref{sec:supervised_coding}. Fig.~\ref{fig:FMD_kSC_dic} depicts the accuracy of the different coding methods as a function of the number of dictionary elements on the FMD dataset. Note that supervised coding consistently improves the classification accuracy for both sparse coding and LLC. Note also that the bilinear classifier outperforms the linear one in most cases.
The best recognition accuracy of $61.8\%$ is achieved by supervised kernel sparse coding using a bilinear classifier. To the best of our knowledge, this constitutes the state-of-the-art performance on FMD (Sharan \etal~\cite{FM_DATABASE} reported $60.6\%$ with feature combination). Note that the best performance reported in~\cite{FM_DATABASE} using only SIFT features is $41.2\%$, which is significantly lower than kernel coding with ultimately the same features.

\def \FMD_SCALE {0.15}
\begin{figure}[!tb]
	\centering
	\includegraphics[width = \FMD_SCALE \textwidth]{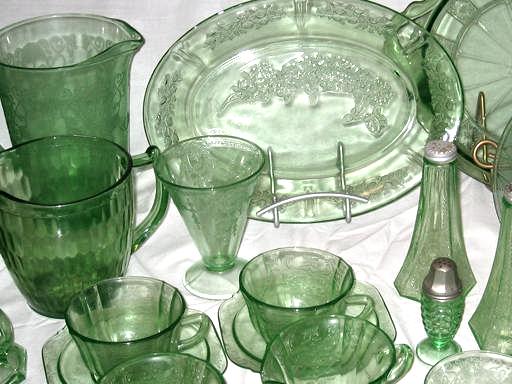}
	\includegraphics[width = \FMD_SCALE \textwidth]{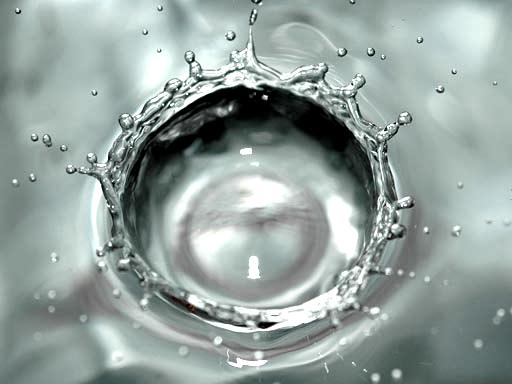}
	\includegraphics[width = \FMD_SCALE \textwidth]{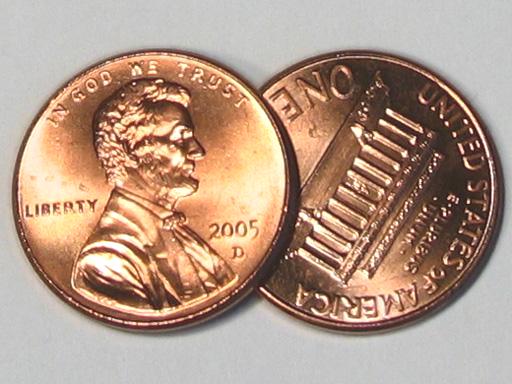}
	\includegraphics[width = \FMD_SCALE \textwidth]{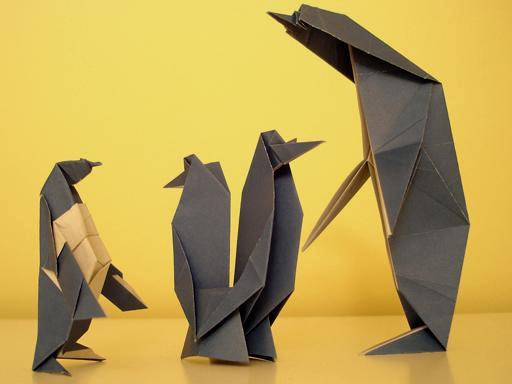}
	\includegraphics[width = \FMD_SCALE \textwidth]{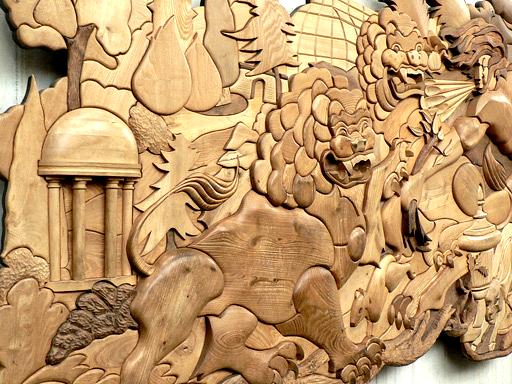}
	\includegraphics[width = \FMD_SCALE \textwidth]{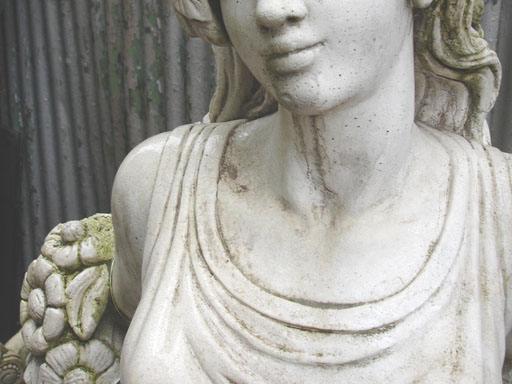}
	\vspace{-0.15cm}
	\caption{\footnotesize Examples from the FMD dataset~\cite{FM_DATABASE}}
	\label{fig:FMD_Dataset}
	\vspace{-0.2cm}
\end{figure}
	
\begin{table}[!tb]
  	\centering
    \begin{tabular*}{1.0\textwidth}{@{\extracolsep{\fill} } lccccc}
    	\toprule {\textbf{Method}}
    	&{SC} &{LLC} &{kBOW} &{kSC} &{kLLC} \\
    	\midrule
    	{\textbf{Accuracy}}
    	&$51.6\% \pm 2.2$ &$48.8\% \pm 1.3$ &$46.1\% \pm 2.0$ &$\bf 53.4\% \pm 1.7$ &$52.8\% \pm 1.6$  \\
    	\bottomrule	
    \end{tabular*}
    \vspace{0.1cm}
 	\caption    {\footnotesize Comparison of different coding techniques on FMD without dictionary learning. }
    \label{tab:table_FMD_performance1}
    \vspace{-0.2cm}
\end{table}

\subsection{MIT Indoor}
As a second experiment, we performed  scene classification using the challenging MIT Indoor dataset~\cite{MIT:Indoor67}.
MIT Indoor contains 67 scene categories with images collected from online sources such as Flicker and Google.
We used a bag of words model with 8000 visual words computed from dense RootSIFT~\cite{Arandjelovic:CVPR:2012} as initial
image descriptor. The final descriptor was obtained by whitening and reducing the dimensionality of the initial descriptor to 4000.
We followed the test protocol used in~\cite{MIT:Indoor67}, where each category contains about 80 training images and 20 test images,
and report the mean Average Precision (mAP).

\begin{figure}[!tb]
	\begin{minipage}{0.45 \textwidth}\centering
		\includegraphics[width = \textwidth]{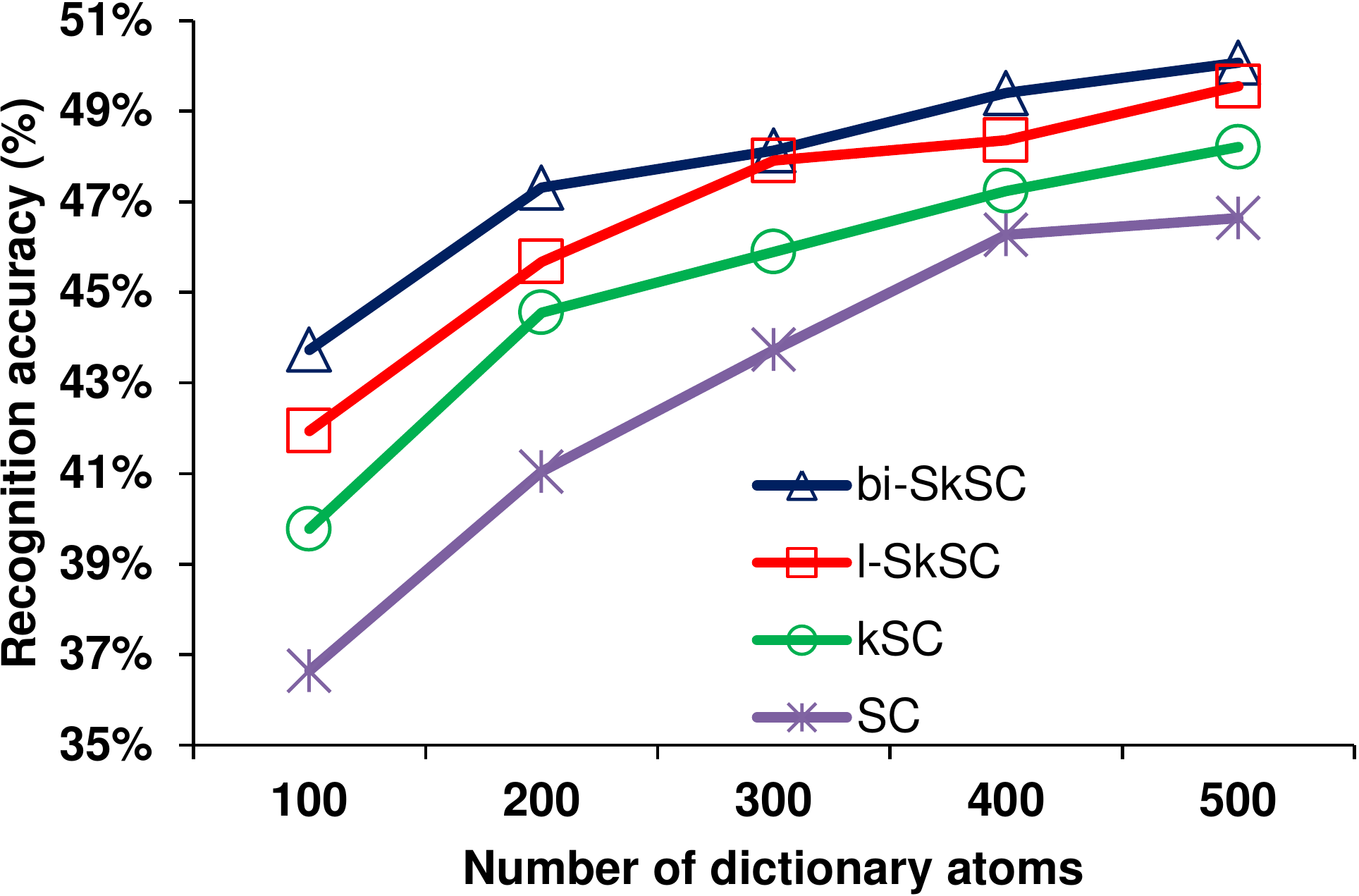}
	\end{minipage}
	\hfill
	\begin{minipage}{0.45 \textwidth}\centering
		\includegraphics[width = \textwidth]{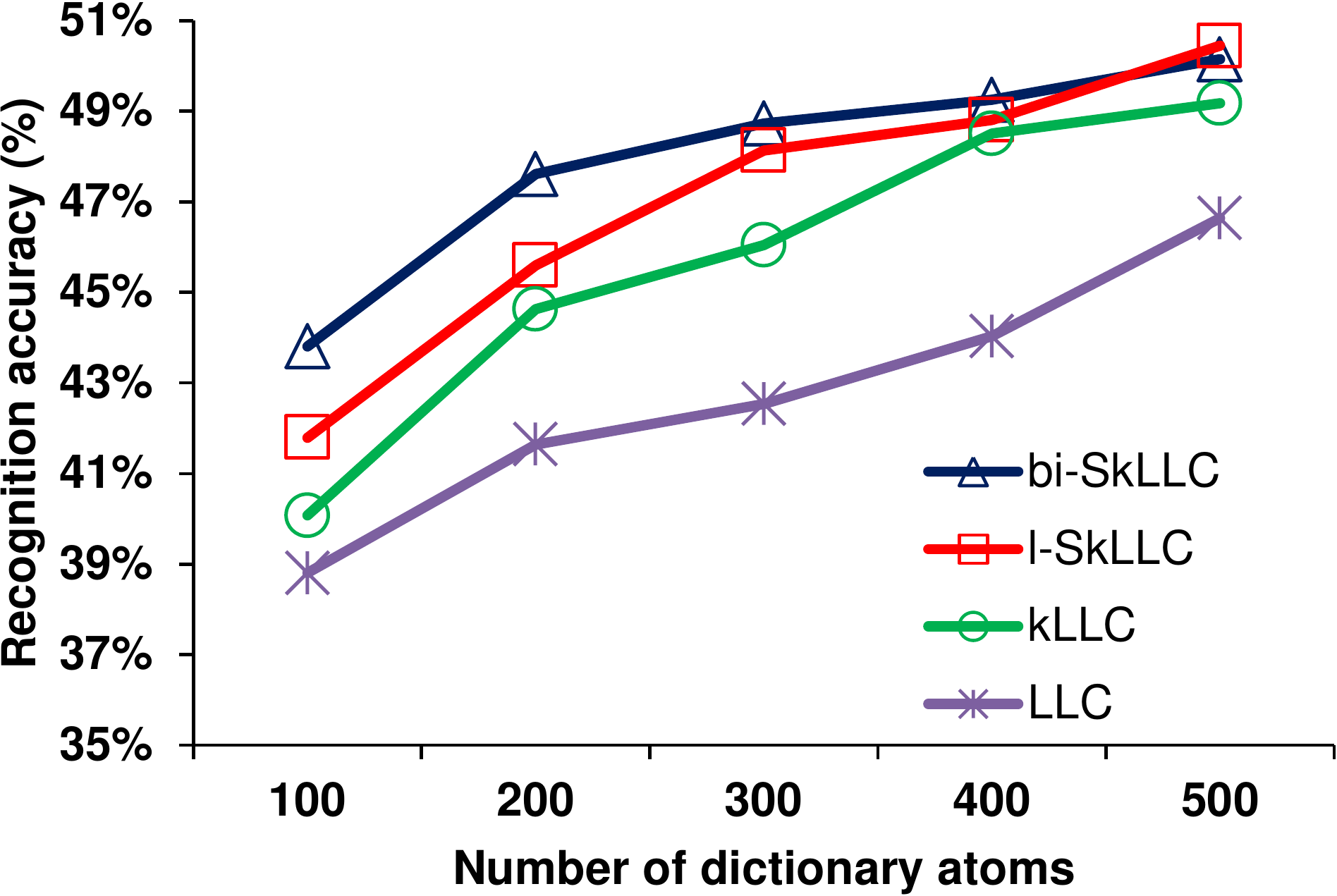}
	\end{minipage}			
	\vspace{-0.2cm}
	\caption{\footnotesize mAP curves for dictionary learning with unsupervised and supervised kernel coding on MIT Indoor.}
	\label{fig:mit_kSC_dic}
	\vspace{-0.2cm}
\end{figure}

In Fig.~\ref{fig:mit_kSC_dic}, we compare the performance of linear coding against kernel coding (supervised and unsupervised).
Several observations can be made from Fig.~\ref{fig:mit_kSC_dic}. First, kernel coding, \ie, kSC and kLLC,
significantly outperform the linear methods, SC and LLC. For example, kLLC with a dictionary of size 500 reaches a mAP of
$49.2\%$ while LLC yields $46.6\%$. Second, supervised learning consistently improves the performance. The maximum
mAP of $50.4\%$ is achieved by l-SkLLC. Importantly, the results obtained here are competitive to several state-of-the-art 
techniques. For example, the mAP reported in~\cite{Singh:ECCV:2012} and ~\cite{Wang:ICML:2013} are of $49.4\%$ and $50.15\%$, respectively. Given the simplicity of the image descriptor employed here, one could expect higher performance if more discriminative features (\eg, Fisher vectors~\cite{Sanchez:IJCV:2013}) were used. This, however, goes beyond the scope of the paper, which rather aims to assess the performance of kernel coding against linear coding.

\subsection{Pascal VOC2007}

As a third experiment, we evaluated the proposed kernel coding methods on the PASCAL Visual Object Classes Challenge (VOC2007)~\cite{Everingham:IJCV:2010}. 
This dataset contains 9963 images from 20 classes, including people, animals, vehicles and indoor objects, and is considered as a realistic and difficult dataset for object classification.
We used the same descriptors as for the MIT Indoor dataset to represent the images.
The dataset is split into training (2501 images), validation (2501 images) and test (4952 images) sets. We used the validation set to obtain the bandwidth of the Gaussian kernels used in our experiments.

\begin{figure}[!tb]
	\begin{minipage}{0.45 \textwidth}\centering
		\includegraphics[width = \textwidth]{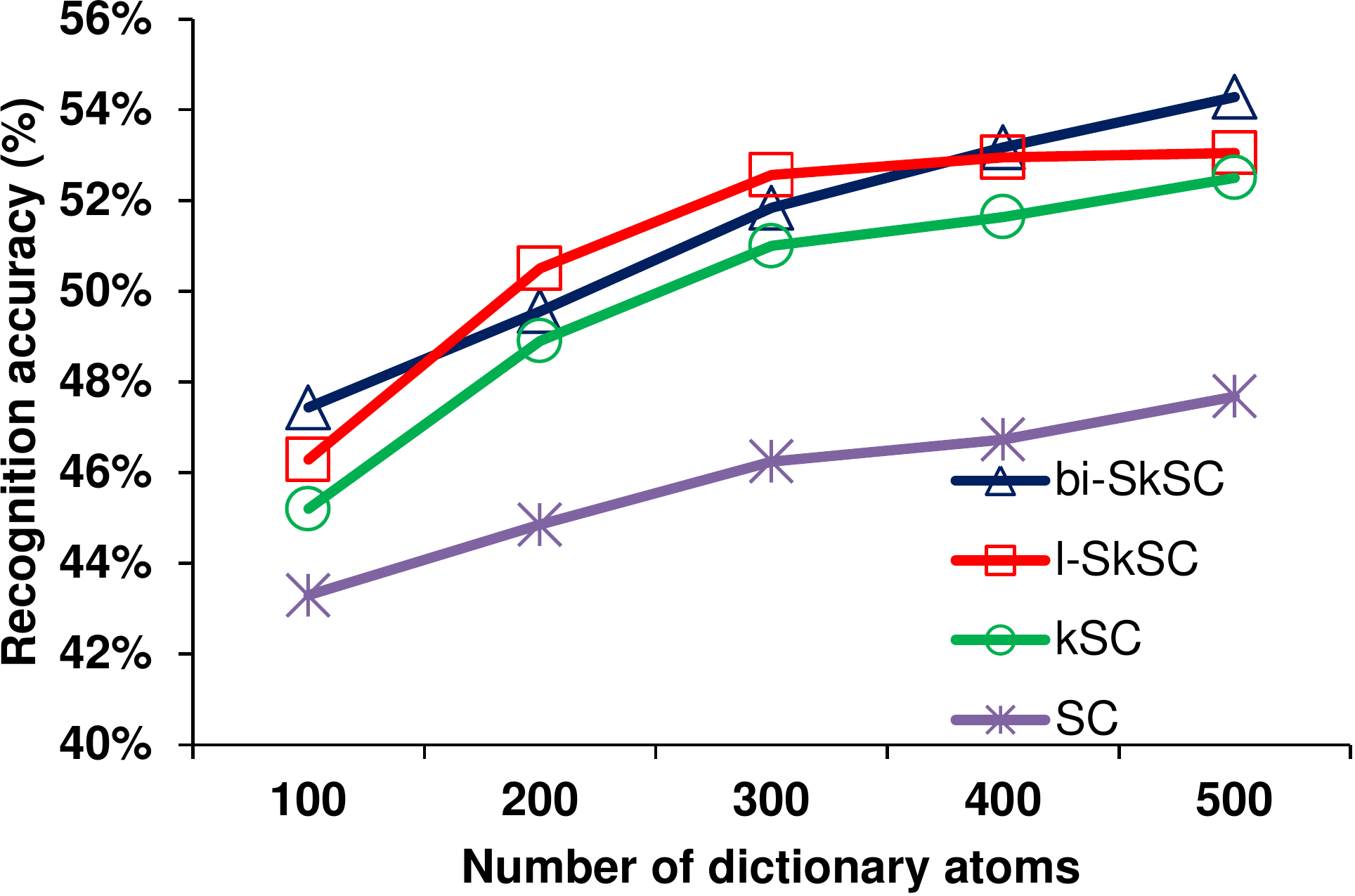}
	\end{minipage}
	\hfill
	\begin{minipage}{0.45 \textwidth}\centering
		\includegraphics[width = \textwidth]{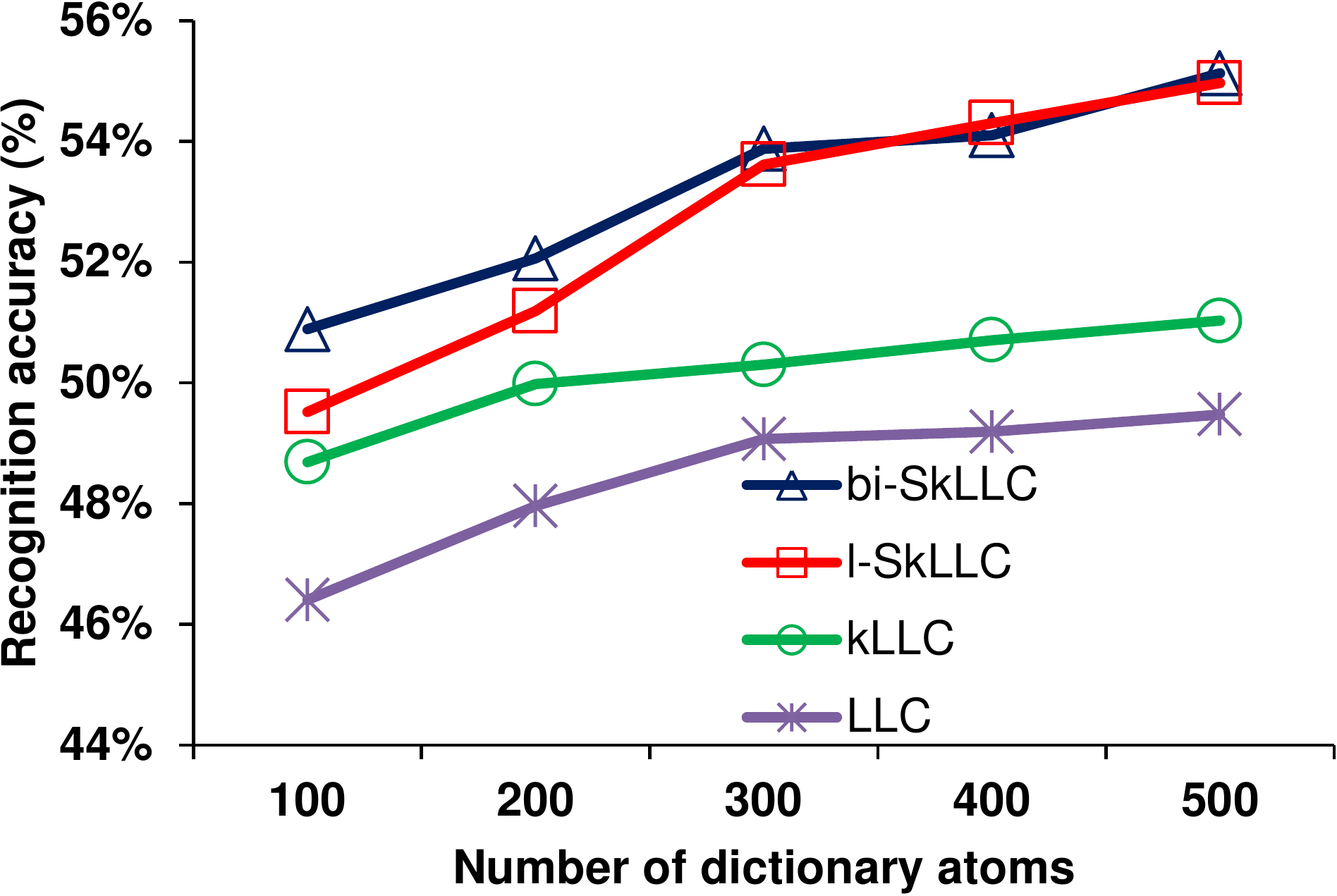}
	\end{minipage}			
	\vspace{-0.2cm}
	\caption{\footnotesize Recognition accuracies of dictionary learning for unsupervised and supervised kernel coding VOC2007.}
	\label{fig:VOC07_kSC_dic}
	\vspace{-0.2cm}
\end{figure}

In Fig.~\ref{fig:VOC07_kSC_dic}, we compare the performance of linear coding against kernel coding (supervised and unsupervised) for VOC2007.
This figure reveals that
\begin{enumerate}
	\item Kernel coding is superior to linear methods for all dictionary sizes. For example, kSC outperforms SC by almost $5\%$ using 500 atoms.
	\item Supervised coding boosts the performance even further. For example, with the linear classifier from~\eqref{eqn:learn_supervised}, the performance of kLLC can be improved from $50.7\%$ to $54.3\%$ with 400
	atoms.
	\item In the majority of cases, the bilinear supervised model outperforms the linear classifier. This, however, is not always the case.
	For example, with 300 atoms, the performance of l-SkSC is $52.6\%$ while bi-SkSC yields $51.8\%$.
\end{enumerate}

\subsection{Extended YaleB}

As a fourth experiment, we evaluated kernel coding against SRC~\cite{Wright:PAMI:2009} and the state-of-the-art LC-kSVD~\cite{Jiang:PAMI:2013}, which can be considered 
as a supervised extension of linear sparse coding and dictionary learning.
To provide a fair comparison, we used the data provided by the authors of~\cite{Jiang:PAMI:2013} for the extended YALE-B dataset~\cite{YALEB_DATASET}.
The extended YaleB database contains 2,414 frontal face images of 38 people (\ie, about 64 images for each person). We used half of the images per category
as training data and the other half for testing purpose using the partition provided by~\cite{Jiang:PAMI:2013}.

We performed two tests on YALEB. In the first one, we evaluated the performance of kSC and kLLC by 
learning a dictionary of size 570 in an unsupervised manner. The results of this test are shown at the top of 
Table~\ref{tab:table_yaleb_performance}. We observe that both kSC and kLLC, without supervised learning,
outperform the state-of-the-art LC-KSVD method. The maximum accuracy of $97.2\%$ is achieved by kLLC which 
outperforms LC-KSVD by more than $2\%$.
For the second test, we considered all training data as dictionary atoms and employed the classification method described 
in Eq.\ref{eqn:sparse_residual_error}. The results of this test
are shown in the bottom part of Table~\ref{tab:table_yaleb_performance}. Again we see that kLLC achieves the highest accuracy of 
$98.4\%$ and thus outperforms LC-KSVD.

\begin{table}[!tb]
  	\centering
    \begin{tabular}{lc}
    	\toprule
    	{\bf Method} &{\bf Recognition Accuracy }\\
    	\toprule  	
    	{\bf SRC-(570 atoms)}							&$80.5\%$\\	
    	{\bf LC-KSVD-(570 atoms)}						&$95.0\%$\\	
    	{\bf kSC-(570 atoms)}							&$ 96.9\%$\\	
    	{\bf kLLC-(570 atoms)}							&$\bf 97.2\%$\\	
    	\midrule
		{\bf SRC-(all training samples)}				&$97.2\%$\\	
    	{\bf LC-KSVD-(all training samples)}			&$96.7\%$\\	 
    	{\bf kSC-(all training samples)}				&$98.2\%$\\	
    	{\bf kLLC-(all training samples)}				&$\bf 98.4\%$\\	    	    			
		\bottomrule	
    \end{tabular}
    \vspace{0.1cm}
 	\caption    {\small Recognition accuracies for the extended YaleB dataset~\cite{YALEB_DATASET}.}
    \label{tab:table_yaleb_performance}
\vspace{-0.3cm}
\end{table}

\subsection{Virus}

Finally, to illustrate the ability of kernel coding methods at handling manifold data, we performed coding on a Riemannian manifold using the virus dataset~\cite{VIRUS_DATASET}.
The virus dataset contains 15 different virus classes. Each class comprises 100 images of size $41 \times 41$ that were automatically segmented~\cite{VIRUS_DATASET}.
We used the 10 splits provided with the dataset in a leave one out manner, \ie, 10 experiments with 9 splits for training and 1 split as query.
In this experiment, we used Region Covariance Matrices (RCM) as image descriptors. To generate the RCM descriptor of an image, at each pixel $(u,v)$ of the image, we computed the 25-dimensional feature vector
\begin{small}
\begin{equation*}
	\Vec{x}_{u,v} =	\hspace{-0.1cm}\bigg[I_{u,v}, \left| \dfrac{\partial I}  {\partial u}  \right|, 
	\left|\frac{\partial I}  {\partial v}  \right|,
    \left| \frac{\partial^2 I}{\partial u^2}\right|, \left|\frac{\partial^2 I}{\partial v^2}\right|, \big|G^{0,0}_{u,v}\big|,
    \cdots,\big|G^{4,5}_{u,v}\big| ~\bigg]^T,
\end{equation*}%
\end{small}
\noindent
where $I_{u,v}$ is the intensity value, $G^{o,s}_{u,v}$ is the response of a 2D Gabor wavelet~\cite{Lee:PAMI:1996}
 with orientation $o$ and scale $s$, and $|\cdot|$ denotes the magnitude of a 
complex value. Here, we generated 20 Gabor filters at 4 orientations and 5 scales. RCM descriptors are symmetric positive definite matrices, and thus lie on a Riemannian manifold where Euclidean methods do not apply. Our framework, however, lets us employ the log-Euclidean RBF kernel recently proposed in~\cite{Sadeep:CVPR:2013}. With this kernel, kSC and kLLC achieved $79.1\%$ and $80.0$, respectively. 
In~\cite{VIRUS_DATASET}, the performance of various features, specifically designed for the task of 
virus classification, was studied. The state-of-the-art accuracy reported in~\cite{VIRUS_DATASET} was $54.5\%$, which both of kernel coding methods clearly outperform. Aside from showing the power of
kernel coding, this experiment also demonstrates that kernel coding can be employed in scenarios where linear coding is not applicable.

\section{Conclusions and Future Work}
In this paper, we have introduced a general formulation for coding in possibly infinite dimensional Reproducing Kernel Hilbert Spaces (RKHS).
Among others, we have studied the kernelization of two popular coding schemes, namely sparse coding and locality-constrained linear coding, and have introduce several extensions such as
supervised coding using linear or bilinear classifiers. Furthermore, we have proposed two ways to identify a discriminative RKHS for  
coding. Our experimental evaluation on several recognition tasks has demonstrated the benefits of the proposed kernel coding schemes over conventional linear coding solutions.
In the future, we plan to investigate the notion of max-margin for supervised kernel coding. 
We are also interested in devising kernel extensions of other coding schemes such as Vector of Locally Aggregated Descriptors (VLAD)~\cite{Jegou:PAMI:2012} and Fisher vectors~\cite{Sanchez:IJCV:2013}.

\end{document}